\def\year{2019}\relax
\documentclass[letterpaper]{article} 
\usepackage{aaai19}  
\usepackage{times}  
\usepackage{helvet}  
\usepackage{courier}  
\usepackage{url}  
\usepackage{graphicx}  
\usepackage{pbox}
\usepackage{xcolor}
\usepackage{booktabs}
\usepackage{amsmath}
\usepackage{multirow}
\usepackage{subfigure}
\usepackage{amssymb}
\usepackage{pifont}
\newcommand{\cmark}{\ding{51}}%
\newcommand{\xmark}{\ding{55}}%

\begin{document}
%
\title{Read + Verify: Machine Reading Comprehension \\ with Unanswerable Questions}
\author{
Minghao Hu$^1$\thanks{Contribution during internship at Microsoft Research Asia.}, 
Furu Wei$^2$, 
Yuxing Peng$^1$, 
Zhen Huang$^1$, 
Nan Yang$^2$,
Dongsheng Li$^1$
\\ 
$^1$College of Computer, National University of Defense Technology  \\
$^2$Microsoft Research Asia   \\
\{huminghao09,pengyuxing,huangzhen,dsli\}@nudt.edu.cn \\
\{fuwei,nanya\}@microsoft.com
}
\maketitle

\begin{abstract}
Machine reading comprehension with unanswerable questions aims to abstain from answering when no answer can be inferred.
In addition to extract answers, previous works usually predict an additional ``no-answer'' probability to detect unanswerable cases.
However, they fail to validate the answerability of the question by verifying the legitimacy of the predicted answer.
To address this problem, we propose a novel read-then-verify system, which not only utilizes a neural reader to extract candidate answers and produce no-answer probabilities, but also leverages an answer verifier to decide whether the predicted answer is entailed by the input snippets.
Moreover, we introduce two auxiliary losses to help the reader better handle answer extraction as well as no-answer detection, and investigate three different architectures for the answer verifier.
Our experiments on the SQuAD 2.0 dataset show that our system obtains a score of 74.2 F1 on test set, achieving state-of-the-art results at the time of submission (Aug. 28th, 2018).
\end{abstract}

\section{Introduction}
The ability to comprehend text and answer questions is crucial for natural language processing.
Due to the creation of various large-scale datasets~\cite{Hermann15,Nguyen16,Joshi17,Kovcisky18}, remarkable advancements have been made in the task of machine reading comprehension.
Nevertheless, one important hypothesis behind current approaches is that there always exists a correct answer in the context passage. 
Therefore, the models only need to choose a most plausible text span based on the question, instead of checking if there exists an answer in the first place.  
Recently, a new version of Stanford Question Answering Dataset (SQuAD), namely SQuAD 2.0~\cite{Rajpurkar18}, has been proposed to test the ability of answering answerable questions as well as detecting unanswerable cases.
To deal with unanswerable cases, systems must learn to identify a wide range of linguistic phenomena such as negation, antonymy and entity changes between the passage and the question.

Previous works~\cite{Levy17,Clark18,kundu2018nil} all apply a shared-normalization operation between a ``no-answer'' score and answer span scores, so as to produce a probability that a question is unanswerable as well as output a candidate answer.
However, they have not considered further validating the answerability of the question by verifying the legitimacy of the predicted answer.
Here, \emph{answerability} denotes whether the question has an answer, and \emph{legitimacy} means whether the extracted text can be supported by the passage and the question.
Human, on the contrary, tends to first find a plausible answer given a question, and then checks if there exists any contradictory semantics.

\begin{figure}
\begin{center}
\includegraphics[width=2.8in]{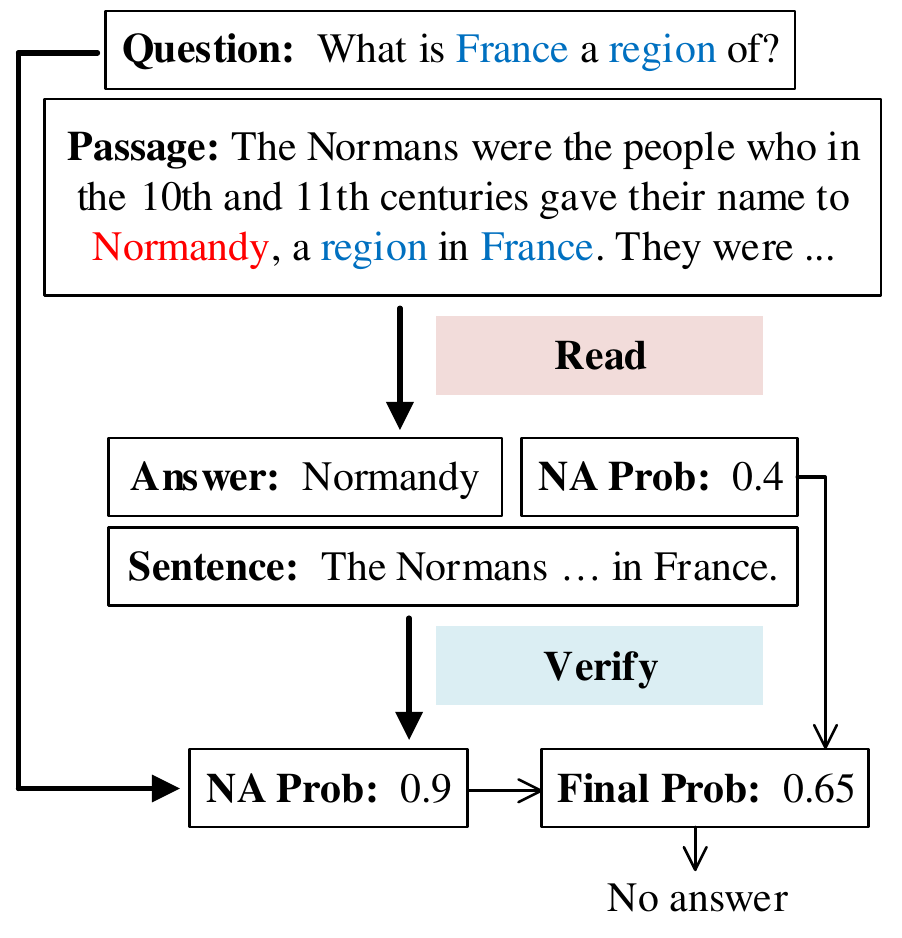}
\vspace{-0.1cm}
\caption{An overview of our approach. 
The reader first extracts a candidate answer and produces a no-answer probability (NA Prob). The answer verifier then checks whether the extracted answer is legitimate or not. Finally, the system aggregates previous results and outputs the final prediction.
}
\label{fig1}
\vspace{-0.5cm}
\end{center}
\end{figure}

To address the above issue, we propose a \emph{read-then-verify} system that aims to be robust to unanswerable questions in this paper.
As shown in Figure \ref{fig1}, our system consists of two components: (1) a \emph{no-answer reader} for extracting candidate answers and detecting unanswerable questions, and (2) an \emph{answer verifier} for deciding whether or not the extracted candidate is legitimate. 
The key contributions of our work are three-fold.

First, we augment existing readers with two \emph{auxiliary losses}, to better handle answer extraction and no-answer detection respectively. 
Since the downstream verifying stage always requires a candidate answer, the reader must be able to extract plausible answers for all questions. 
However, previous approaches are not trained to find potential candidates for unanswerable questions.
We solve this problem by introducing an independent span loss that aims to concentrate on the answer extraction task regardless of the answerability of the question.
In order to not conflict with no-answer detection, we leverage a \emph{multi-head pointer network} to generate two pairs of span scores, where one pair is normalized with the no-answer score and the other is used for our auxiliary loss.
Besides, we present another independent no-answer loss to further alleviate the confliction, by focusing on the no-answer detection task without considering the shared normalization of answer extraction.

Second, in addition to the standard reading phase, we introduce an additional answer verifying phase, which aims at finding local entailment that supports the answer by comparing the answer sentence with the question.
This is based on the observation that the core phenomenon of unanswerable questions usually occurs between a few passage words and question words.
Take Figure \ref{fig1} for example, after comparing the passage snippet ``\emph{Normandy, a region in France}'' with the question, we can easily determine that no answer exists since the question asks for an \emph{impossible condition}\footnote{Impossible condition means that the question asks for something that is not satisfied by anything in the given passage.}.
This observation is even more obvious when \emph{antonym} or \emph{mutual exclusion} occurs, such as the question asks for ``\emph{the decline of rainforests}'' but the passage mentions that ``\emph{the rainforests spread out}''.
Inspired by recent advances in natural language inference (NLI)~\cite{Bowman15}, we investigate three different architectures for the answer verifying task.
The first one is a sequential model that takes two sentences as a long sequence, while the second one attempts to capture interactions between two sentences.
The last one is a hybrid model that combines the above two models to test if the performance can be further improved.

Lastly, we evaluate our system on the SQuAD 2.0 dataset~\cite{Rajpurkar18}, a reading comprehension benchmark augmented with unanswerable questions. Our best reader achieves a F1 score of 73.7 and 69.1 on the development set, with or without ELMo embeddings~\cite{Elmo17}.
When combined with the answer verifier, the whole system improves to 74.8 F1 and 71.5 F1 respectively.
Moreover, the best system obtains a score of 74.2 F1 on test set, achieving state-of-the-art results at the time of submission (Aug. 28th, 2018).

\section{Background}
Existing reading comprehension models focus on answering questions where a correct answer is guaranteed to exist.
However, they are not able to identify unanswerable questions but tend to return an unreliable text span.
Consequently, we first give a brief introduction on the unanswerable reading comprehension task, and then investigate current solutions.

\subsection{Task Description}
Given a context passage and a question, the machine needs to not only find answers to answerable questions but also detect unanswerable cases. 
The passage and the question are described as sequences of word tokens, denoted as $P=\{x_i^p\}_{i=1}^{l_p}$ and $Q=\{x_j^q\}_{j=1}^{l_q}$ respectively, where $l_p$ is the passage length and $l_q$ is the question length. 
Our goal is to predict an answer $A$, which is constrained as a segment of text in the passage: $A=\{x_i^p\}_{i=l_a}^{l_b}$, or return an empty string if there is no answer, where $l_a$ and $l_b$ indicate the answer boundary. 

\subsection{No-Answer Reader}
To predict an answer span, current approaches first embed and encode both of passage and question into two series of fix-sized vectors. Then they leverage various attention mechanisms, such as bi-attention~\cite{Seo17} or reattention~\cite{Hu17}, to build interdependent representations for passage and question, which are denoted as $U=\{u_i\}_{i=1}^{l_p}$ and $V=\{v_j\}_{j=1}^{l_q}$ respectively.
Finally, they summarize the question representation into a dense vector $t$, and utilize the pointer network~\cite{Vinyals15} to produce two scores over passage words that indicate the answer boundary~\cite{Wang17b}:
\begin{gather}
	o_{j} = w_v^ \mathrm{ T } v_j  \ , \
	t=\sum_{j=1}^{l_q} \frac{e^{o_{j}}}{\sum_{k=1}^{l_q} e^{o_{k}}} v_j \nonumber \\
    \alpha, \beta = \mathrm{pointer\_network}(U, t)	\nonumber
\end{gather}
where $\alpha$ and $\beta$ are the \emph{span scores} for answer start and end bounds.

In order to additionally detect if the question is unanswerable, previous approaches~\cite{Levy17,Clark18,kundu2018nil} attempt to predict a special \emph{no-answer score} $z$ in addition to the distribution over answer spans.
Concretely, a shared softmax function can be applied to normalize both of no-answer score and span scores, yielding a joint no-answer objective defined as:
\begin{eqnarray} 
	\mathcal{L}_{joint} = - \log \left( \frac{(1 - \delta)e^z + \delta e^{\alpha_a \beta_b} }{e^z + \sum_{i=1}^{l_p} \sum_{j=1}^{l_p} e^{\alpha_i \beta_j}} \right)	\nonumber
\end{eqnarray}
where $a$ and $b$ are the ground-truth start and end positions, and $\delta$ is 1 if the question is answerable and 0 otherwise.
At test time, a question is detected as being unanswerable once the normalized no-answer score exceeds some threshold.

\section{Approach}
In this section we describe our proposed read-then-verify system. 
The system first leverages a neural reader to extract a candidate answer and detect if the question is unanswerable.
It then utilizes an answer verifier to further check the legitimacy of the predicted answer.
We enhance the reader with two novel auxiliary losses, and investigate three different architectures for the answer verifier.

\begin{figure*}
\begin{center}
\includegraphics[width=6.8in]{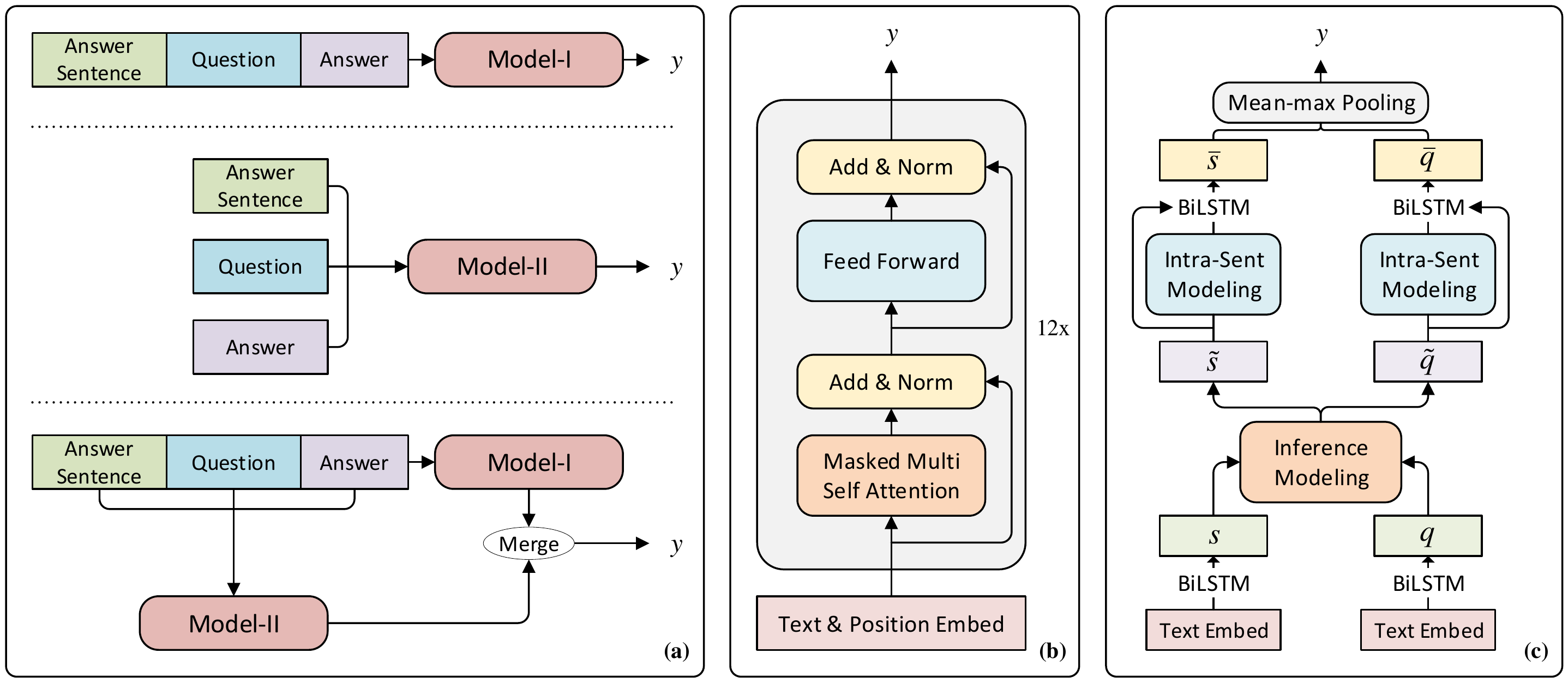}
\vspace{-0.1cm}
\caption{An overview of answer verifiers. 
(a) Input structures for running three different models.
(b) Generative Pre-trained Transformer proposed by Radford et al.~\shortcite{Radford18}. Here, ``Masked Multi Self Attention'' refers to multi-head self-attention function~\cite{vaswani2017attention} that only attends to previous tokens. ``Add \& Norm'' indicates residual connection and layer normalization.
(c) Our proposed token-wise interaction model, which is designed to compare two sentences and aggregate the results for verifying the answer.
}
\label{fig2}
\vspace{-0.5cm}
\end{center}
\end{figure*}

\subsection{Reader with Auxiliary Losses}
Although previous no-answer readers are capable of jointly learning answer extraction and no-answer detection, there exists two problems for each individual task.
For the answer extraction, previous readers are not trained to find candidate answers for unanswerable questions.
In our system, however, the reader is required to extract a plausible answer that is fed to the downstream verifying stage for all questions.
As for no-answer detection, a \emph{confliction} could be triggered due to the shared normalization between span scores and no-answer score.
Since the sum of these normalized scores is always 1, an over-confident span probability would cause an unconfident no-answer probability, and vice versa.
Therefore, inaccurate confidence on answer span, which has been observed by Clark et al.~\shortcite{Clark18}, could lead to imprecise prediction on no-answer score.
To address the above issues, we propose two auxiliary losses to optimize and enhance each task independently without interfering with each other.

\subsubsection{Independent Span Loss}
This loss is designed to concentrate on answer extraction.  
In this task, the model is asked to extract candidate answers for all possible questions. 
Therefore, besides answerable questions, we also include unanswerable cases as positive examples, and consider the \emph{plausible answer} as gold answer\footnote{In SQuAD 2.0, the plausible answer is annotated by human for every unanswerable question. A pre-trained reader can also be used to extract plausible answers if no annotation is provided.}.
In order to not conflict with no-answer detection, we propose to use a multi-head pointer network to additionally produce another pair of span scores $\tilde{\alpha}$ and $\tilde{\beta}$:
\begin{gather}
	\tilde{o}_{j} = \tilde{w}_v^ \mathrm{ T } v_j  \ , \
	\tilde{t}=\sum_{j=1}^{l_q} \frac{e^{\tilde{o}_{j}}}{\sum_{k=1}^{l_q} e^{\tilde{o}_{k}}} v_j \nonumber \\
    \tilde{\alpha}, \tilde{\beta} = \mathrm{pointer\_network}(U, \tilde{t})	\nonumber
\end{gather}
where multiple heads share the same network architecture but with different parameters. 

Then, we define an independent span loss as:
\begin{eqnarray} 
	\mathcal{L}_{indep-\uppercase\expandafter{\romannumeral1}} = - \log \left( \frac{ e^{ \tilde{\alpha}_{\tilde{a}} \tilde{\beta}_{\tilde{b}} } }{\sum_{i=1}^{l_p} \sum_{j=1}^{l_p} e^{\tilde{\alpha}_i \tilde{\beta}_j}} \right)	\nonumber
\end{eqnarray}
where $\tilde{a}$ and $\tilde{b}$ are the augmented ground-truth answer boundaries. The final span probability is obtained using a simple mean pooling over the two pairs of softmax-normalized span scores.

\subsubsection{Independent No-Answer Loss}
Despite a multi-head pointer network being used to prevent the confliction problem, no-answer detection can still be weakened since the no-answer score $z$ is normalized with span scores.
Therefore, we consider exclusively encouraging the prediction on no-answer detection.
This is achieved by introducing an independent no-answer loss as:
\begin{eqnarray} 
	\mathcal{L}_{indep-\uppercase\expandafter{\romannumeral2}} = - (1 - \delta) \log \sigma(z) - \delta \log(1 - \sigma(z))	\nonumber
\end{eqnarray}
where $\sigma$ is the sigmoid activation function. 
Through this loss, we expect the model to produce a more confident prediction on no-answer score $z$ without considering the shared-normalization operation.

Finally, we combine the above losses as follows:
\begin{eqnarray} 
	\mathcal{L} = \mathcal{L}_{joint} + \gamma \mathcal{L}_{indep-\uppercase\expandafter{\romannumeral1}} + \lambda \mathcal{L}_{indep-\uppercase\expandafter{\romannumeral2}}	\nonumber
\end{eqnarray}
where $\gamma$ and $\lambda$ are two hyper-parameters that control the weight of two auxiliary losses.

\subsection{Answer Verifier}
After the answer is extracted, an answer verifier is used to compare the answer sentence with the question, so as to recognize local textual entailment that supports the answer.
Here, we define the \emph{answer sentence} as the context sentence that contains either gold answers or plausible answers.
We explore three different architectures, as shown in Figure \ref{fig2}: 
(1) a sequential model that takes the inputs as a long sequence, (2) an interactive model that encodes two sentences interdependently, and (3) a hybrid model that takes both of the two approaches into account.

\subsubsection{Model-\uppercase\expandafter{\romannumeral1}: Sequential Architecture}
In Model-\uppercase\expandafter{\romannumeral1}, we convert the answer sentence and the question along with the extracted answer into an ordered input sequence. 
Then we adapt the recently proposed Generative Pre-trained Transformer (OpenAI GPT)~\cite{Radford18} to perform the task. 
The model is a multi-layer Transformer decoder~\cite{Liu18}, which is first trained with a language modeling objective on a large unlabeled text corpus and then finetuned on the specific target task.

Specifically, given an answer sentence $S$, a question $Q$ and an extracted answer $A$, we concatenate the two sentences with the answer while adding a delimiter token in between to get $[S; Q; \$; A]$. 
We then embed the sequence with its word embedding as well as position embedding.
Multiple transformer blocks are used to encode the sequence embeddings as follows:
\begin{gather}
    h_0 = W_e[X] + W_p  \nonumber \\
    h_i = \mathrm{transformer\_block}(h_{i-1}), \forall i \in [1,n]	\nonumber
\end{gather}
where $X$ denotes the sequence's indexes in the vocab, $W_e$ is the token embedding matrix, $W_p$ is the position embedding matrix, and $n$ is the number of transformer blocks. 
Each block consists of a masked multi-head self-attention layer~\cite{vaswani2017attention} and a position-wise feed-forward layer. Residual connection and layer normalization are used after each layer.

The last token's activation $h_n^{l_m}$ is then fed into a linear projection layer followed by a $\mathrm{softmax}$ function to output the no-answer probability $y$:
\begin{eqnarray} 
	p(y|X) = \mathrm{softmax}(h_n^{l_m} W_y)	\nonumber
\end{eqnarray}

A standard cross-entropy objective is used to minimize the negative log-likelihood:
\begin{eqnarray} 
	\mathcal{L}(\theta) = - \sum_{(X,y)} \log p(y|X)	\nonumber
\end{eqnarray}

\subsubsection{Model-\uppercase\expandafter{\romannumeral2}: Interactive Architecture}
In Model-\uppercase\expandafter{\romannumeral2}, we consider an interactive architecture that aims to capture the interactions between two sentences, so as to recognize their local entailment relationships for verifying the answer. This model consists of the following layers:

\noindent{\bf Encoding:} 
We embed words using the GloVe embedding~\cite{Pennington14}, and also embed characters of each word with trainable vectors. 
We run a bidirectional LSTM (BiLSTM)~\cite{Hochreiter97} to encode the characters and concatenate two last hidden states to get character-level embeddings. 
In addition, we use a binary feature to indicate if a word is part of the answer.
All embeddings along with the feature are then concatenated and encoded by a weight-shared BiLSTM, yielding two series of contextual representations:
\begin{gather}
	s_i=\mathrm{BiLSTM}([\mathrm{word}_i^s; \mathrm{char}_i^s; \mathrm{fea}_i^s]), \forall i \in [1,l_s] \nonumber \\
	q_j=\mathrm{BiLSTM}([\mathrm{word}_j^q; \mathrm{char}_j^q; \mathrm{fea}_j^q]), \forall j \in [1,l_q] \nonumber  
\end{gather}
where $l_s$ is the length of answer sentence, and $[\cdot;\cdot]$ denotes concatenation.

\noindent{\bf Inference Modeling:} 
An inference modeling layer is used to capture the interactions between two sentences and produce two inference-aware sentence representations. 
We first compute the dot products of all tuples $<s_i, q_j>$ as attention weights, and then normalize these weights so as to obtain attended vectors as follows:
\begin{gather}
	a_{ij} = s_i ^ \mathsf{T} q_j, \forall i \in [1,l_s], \forall j \in [1,l_q] \nonumber \\
	b_i = \sum_{j=1}^{l_q} \frac{e^{a_{ij}}}{\sum_{k=1}^{l_q} e^{a_{ik}}} q_j \ , \
	c_j = \sum_{i=1}^{l_s} \frac{e^{a_{ij}}}{\sum_{k=1}^{l_s} e^{a_{kj}}} s_i \nonumber
\end{gather}
Here, $b_i$ refers to the attended vector from question $Q$ for the $i$-th word in answer sentence $S$, and vice versa for $c_j$.

Next, in order to separately compare the aligned pairs $\{(s_i, b_i)\}_{i=1}^{l_s}$ and $\{(q_j, c_j)\}_{j=1}^{l_q}$ for finding local inference information, we use a weight-shared function $F$ to model these aligned pairs as:
\begin{gather}
	\tilde{s}_i = F(s_i, b_i) \ , \
	\tilde{q}_j = F(q_j, c_j) \nonumber
\end{gather}
$F$ can have various forms, such as BiLSTM, multilayer perceptron, and so on. 
Here we use a heuristic function $o=F(x,y)$ proposed by Hu et al.~\shortcite{Hu17}, which demonstrates good performances compared to other options:
\begin{gather}
r = \mathrm{gelu} \left(W_r [x; y; x \circ y; x - y] \right) \nonumber \\
g = \sigma \left(W_g [x; y; x \circ y; x - y] \right) \nonumber \\
o = g \circ r  + (1 - g) \circ x \nonumber
\end{gather}
where $\mathrm{gelu}$ is the Gaussian Error Linear Unit~\cite{Hendrycks16}, $\circ$ is element-wise multiplication, and the bias term is omitted.

\noindent{\bf Intra-Sentence Modeling:}
Next we apply an intra-sentence modeling layer to capture self correlations inside each sentence.
The input are inference-aware vectors $\tilde{s}_i$ and $\tilde{q}_j$, which are first passed through another BiLSTM layer for encoding.
We then use the same attention mechanism described above, only now between each sentence and itself, and we set $a_{ij}=-inf$ if $i=j$ to ensure that the word is not aligned with itself. 
Another function $F$ is used to produce self-aware vectors $\hat{s}_i$ and $\hat{q}_j$ respectively.

\noindent{\bf Prediction:} 
Before the final prediction, we apply a concatenated residual connection and model the sentences with a BiLSTM as:
\begin{eqnarray} 
	\bar{s}_i=\mathrm{BiLSTM}([\tilde{s}_i;\hat{s}_i])	\ , \ 
	\bar{q}_j=\mathrm{BiLSTM}([\tilde{q}_j;\hat{q}_j])   \nonumber
\end{eqnarray}

A mean-max pooling operation is then applied to summarize the final representation of two sentences, namely $\bar{s}_i$ and $\bar{q}_j$.
All summarized vectors are then concatenated and fed into a feed-forward classifier that consists of a projection sublayer with $\mathrm{gelu}$ activation and a $\mathrm{softmax}$ output sublayer, yielding the no-answer probability. 
As before, we optimize the negative log-likelihood objective function.

\subsubsection{Model-\uppercase\expandafter{\romannumeral3}: Hybrid Architecture}
To explore how the features extracted by Model-\uppercase\expandafter{\romannumeral1} and Model-\uppercase\expandafter{\romannumeral2} can be integrated to obtain better representation capacities, we investigate the combination of the above two models, namely Model-\uppercase\expandafter{\romannumeral3}. 
We merge the output vectors of two models into a single joint representation. 
An unified feed-forward classifier is then applied to output the no-answer probability.
Such design allows us to test whether the performance can benefit from the integration of two different architectures. 
In practice we use a simple concatenation to merge the two sources of information.

\section{Experimental Setup}

\subsection{Dataset}
We evaluate our approach on the SQuAD 2.0 dataset~\cite{Rajpurkar18}. 
SQuAD 2.0 is a new machine reading comprehension benchmark that aims to test the models whether they have truely understood the questions by knowing what they don't know. 
It combines answerable questions from the previous SQuAD 1.1 dataset~\cite{Rajpurkar16} with 53,775 unanswerable questions about the same passages.
Crowdsourcing workers craft these questions with a plausible answer in mind, and make sure that they are relevant to the corresponding passages.

\subsection{Training and Inference}
Our no-answer reader is trained on context passages, while the answer verifier is trained on oracle answer sentences.
Model-\uppercase\expandafter{\romannumeral1} follows a procedure of unsupervised pre-training and supervised fine-tuning. 
That is, the model is first optimized with a language modeling objective on a large unlabeled text corpus to initialize its parameters. 
Then it adapts the parameters to the answer verifying task with our supervised objective.
For Model-\uppercase\expandafter{\romannumeral2}, we directly train it with the supervised loss.
Model-\uppercase\expandafter{\romannumeral3}, however, consists of two different architectures that require different training procedures. Therefore, we initialize Model-\uppercase\expandafter{\romannumeral3} with the pre-trained parameters from both of Model-\uppercase\expandafter{\romannumeral1} and Model-\uppercase\expandafter{\romannumeral2}, and then fine-tune the whole model until convergence.

At test time, the reader first predicts a candidate answer as well as a passage-level no-answer probability. 
The answer verifier then validates the extracted answer along with its sentence and outputs a sentence-level probability. 
Following the official evaluation setting, a question is detected to be unanswerable once the joint no-answer probability, which is computed as the mean of the above two probabilities, exceeds some threshold.
We tune this threshold to maximize F1 score on the development set, and report both of EM (Exact Match) and F1 metrics.
We also evaluate the performance on no-answer detection with an accuracy metric (ACC), where its threshold is set as 0.5 by default.

\subsection{Implementation}
We use the Reinforced Mnemonic Reader (RMR)~\cite{Hu17}, one of the state-of-the-art reading comprehension models on the SQuAD 1.1 dataset, as our base reader.
The reader is configurated with its default setting, and trained with the no-answer objective with our auxiliary losses.
ELMo (Embeddings from Language Models)~\cite{Elmo17} is exclusively listed in our experimental configuration.
We run a grid search on $\gamma$ and $\lambda$ among [0.1, 0.3, 0.5, 0.7, 1, 2]. 
Based on the performance on development set, we set $\gamma$ as 0.3 and $\lambda$ to be 1.
As for answer verifiers, we use the original configuration from Radford et al.~\shortcite{Radford18} for Model-\uppercase\expandafter{\romannumeral1}.
For Model-\uppercase\expandafter{\romannumeral2}, the Adam optimizer~\cite{Kingma14} with a learning rate of 0.0008 is used, the hidden size is set as 300, and a dropout~\cite{Srivastava14} of 0.3 is applied for preventing overfitting.
The batch size is 48 for the reader, 64 for Model-\uppercase\expandafter{\romannumeral2}, and 32 for Model-\uppercase\expandafter{\romannumeral1} as well as Model-\uppercase\expandafter{\romannumeral3}. 
We use the GloVe~\cite{Pennington14} 100D embeddings for the reader, and 300D embeddings for Model-\uppercase\expandafter{\romannumeral2} and Model-\uppercase\expandafter{\romannumeral3}.
We utilize the \emph{nltk} tokenizer\footnote{https://www.nltk.org/} to preprocess passages and questions, as well as split sentences.
The passages and the sentences are truncated to not exceed 300 words and 150 words respectively.
\section{Evaluation}

\begin{table}
\begin{center}
\begin{tabular}{lcccc}
\toprule
\multirow{2}*{ Model } & \multicolumn{2}{c}{ Dev } & \multicolumn{2}{c}{ Test } \\
 & EM & F1 & EM & F1 \\ 
\midrule
BNA$^1$                        & 59.8 & 62.6 & 59.2 & 62.1 \\ 
DocQA$^2$                      & 61.9 & 64.8 & 59.3 & 62.3 \\
DocQA + ELMo                   & 65.1 & 67.6 & 63.4 & 66.3 \\
ARRR$^\dagger$                  & - & - & 68.6 & 71.1 \\
VS$^3-$Net$^\dagger$           & - & - & 68.4 & 71.3 \\
SAN$^3$                        & - & - & 68.6 & 71.4 \\
FusionNet++(ensemble)$^4$      & - & - & 70.3 & 72.6 \\
SLQA+$^5$ 					   & - & - & 71.5 & \bf{74.4} \\
RMR + ELMo + Verifier          & \bf{72.3} & \bf{74.8} & \bf{71.7} & 74.2 \\
\midrule
Human                          & 86.3 & 89.0 & 86.9 & 89.5 \\
\bottomrule
\end{tabular}
\caption{\label{table1} Comparison of different approaches on the SQuAD 2.0 test set, extracted on Aug 28, 2018: Levy et al.\protect~\shortcite{Levy17}$^1$, Clark et al.\protect~\shortcite{Clark18}$^2$, Liu et al.\protect~\shortcite{liu2017stochastic}$^3$, Huang et al.\protect~\shortcite{Huang17b}$^4$ and Wang et al.\protect~\shortcite{wang2018multi}$^5$. $\dagger$ indicates unpublished works.}
\vspace{-0.3cm}
\end{center}
\end{table}

\subsection{Main Results}
We first submit our approach on the hidden test set of SQuAD 2.0 for evaluation, which is shown in Table \ref{table1}. 
We use Model-\uppercase\expandafter{\romannumeral3} as the default answer verifier, and only report the best result.
As we can see, our system obtains state-of-the-art results by achieving an EM score of 71.7 and a F1 score of 74.2 on the test set. 
Notice that SLQA+ has reached a comparable result compared to our approach. 
We argue that its promising result is largely due to its superior performance compared to our base reader\footnote{SLQA+ achieves 87.0 F1 on the SQuAD 1.1 test set, while RMR reaches 86.6.}.

\subsection{Ablation Study}
Next, we do an ablation study on the SQuAD 2.0 development set to show the effects of our proposed methods for each individual component.
Table \ref{table2} first shows the ablation results of different auxiliary losses on the reader.
Removing the independent span loss (indep-\uppercase\expandafter{\romannumeral1}) results in a performance drop for all answerable questions (HasAns), indicating that this loss helps the model in better identifying the answer boundary.
Ablating independent no-answer loss (indep-\uppercase\expandafter{\romannumeral2}), on the other hand, causes little influence on HasAns, but leads to a severe decline on no-answer accuracy (NoAns ACC).
This suggests that a confliction between answer extraction and no-answer detection indeed happens.
Finally, deleting both of two losses causes a degradation of more than 1.5 points on the overall performance in terms of F1, with or without ELMo embeddings.

Table \ref{table3} details the results of various architectures for the answer verifier.
Model-\uppercase\expandafter{\romannumeral3} outperforms all of other competitors, achieving a no-answer accuracy of 76.2. 
This illustrates that the combination of two different architectures can bring in further improvement.
Adding ELMo embeddings, however, does not boost the performance. 
We hythosize that the bytepair encoding~\cite{Sennrich16} from Model-\uppercase\expandafter{\romannumeral1} and the word/character embeddings from Model-\uppercase\expandafter{\romannumeral2} have provided enough representation capacities.

\begin{table}
\begin{center}
\begin{tabular}{l|ccccc}
\toprule
\multirow{2}*{ Configuration } & \multicolumn{2}{c}{ HasAns } & \multicolumn{2}{c}{ All } & NoAns \\
& EM & F1 & EM & F1 & ACC \\ 
\midrule
RMR    & {72.6} & {81.6} & {66.9} & {69.1} & {73.1} \\
\ \ - indep-\uppercase\expandafter{\romannumeral1}  & \underline{71.3} & \underline{80.4} & 66.0 & 68.6 & 72.8 \\
\ \ - indep-\uppercase\expandafter{\romannumeral2}  & 72.4 & 81.4 & \underline{64.0} & \underline{66.1} & \underline{69.8} \\
\ \ - both    & 71.9 & 80.9 & 65.2 & 67.5 & 71.4 \\
\midrule
RMR + ELMo    & {79.4} & {86.8} & {71.4} & {73.7} & {77.0} \\
\ \ - indep-\uppercase\expandafter{\romannumeral1}  & 78.9 & 86.5 & 71.2 & 73.5 & 76.7 \\
\ \ - indep-\uppercase\expandafter{\romannumeral2}  & 79.5 & 86.6 & \underline{69.4} & \underline{71.4} & \underline{75.1} \\
\ \ - both                                          & \underline{78.7} & \underline{86.2} & 70.0 & 71.9 & 75.3 \\
\bottomrule
\end{tabular}
\caption{\label{table2} Comparison of readers with different auxiliary losses.}
\vspace{-0.1cm}
\end{center}
\end{table}

\begin{table}
\begin{center}
\begin{tabular}{l|cc}
\toprule
Configuration & NoAns ACC\\ 
\midrule
Model-\uppercase\expandafter{\romannumeral1}            & 74.5 \\
Model-\uppercase\expandafter{\romannumeral2}            & 74.6 \\
Model-\uppercase\expandafter{\romannumeral2} + ELMo     & 75.3 \\
Model-\uppercase\expandafter{\romannumeral3}            & \bf{76.2} \\
Model-\uppercase\expandafter{\romannumeral3} + ELMo     & 76.1 \\
\bottomrule
\end{tabular}  
\caption{\label{table3} Comparison of different architectures for the answer verifier.}
\vspace{-0.5cm}
\end{center}
\end{table}

After doing separate ablations on each component, we then compare the performance of the whole system, as shown in Table \ref{table4}.
The combination of base reader with any answer verifier can always result in considerable performance gains, and combining the reader with Model-\uppercase\expandafter{\romannumeral3} obtains the best result.
We find that the improvement on no-answer accuracy is significant. 
This metric raises from 73.1 to 77.1 after adding Model-\uppercase\expandafter{\romannumeral3} to RMR, increasing by 4 absolute points.
Similar observation can be found when ELMo embeddings are used, demonstrating that the gains are consistent and stable.

In order to investigate how the readers affect the overall performance, we fix the answer verifier as Model-\uppercase\expandafter{\romannumeral3} and use DocQA~\cite{Clark18} as the base reader instead of RMR, as shown in Table \ref{table5}. 
We find that the absolute improvements are even larger: the no-answer accuracy roughly increases by 6 points when adding Model-\uppercase\expandafter{\romannumeral3} to DocQA (from 69.1 to 75.2), and 5.5 points when adding Model-\uppercase\expandafter{\romannumeral3} to DocQA + ELMo (from 70.6 to 76.1).

\begin{figure}
\begin{center}
\includegraphics[width=2.9in]{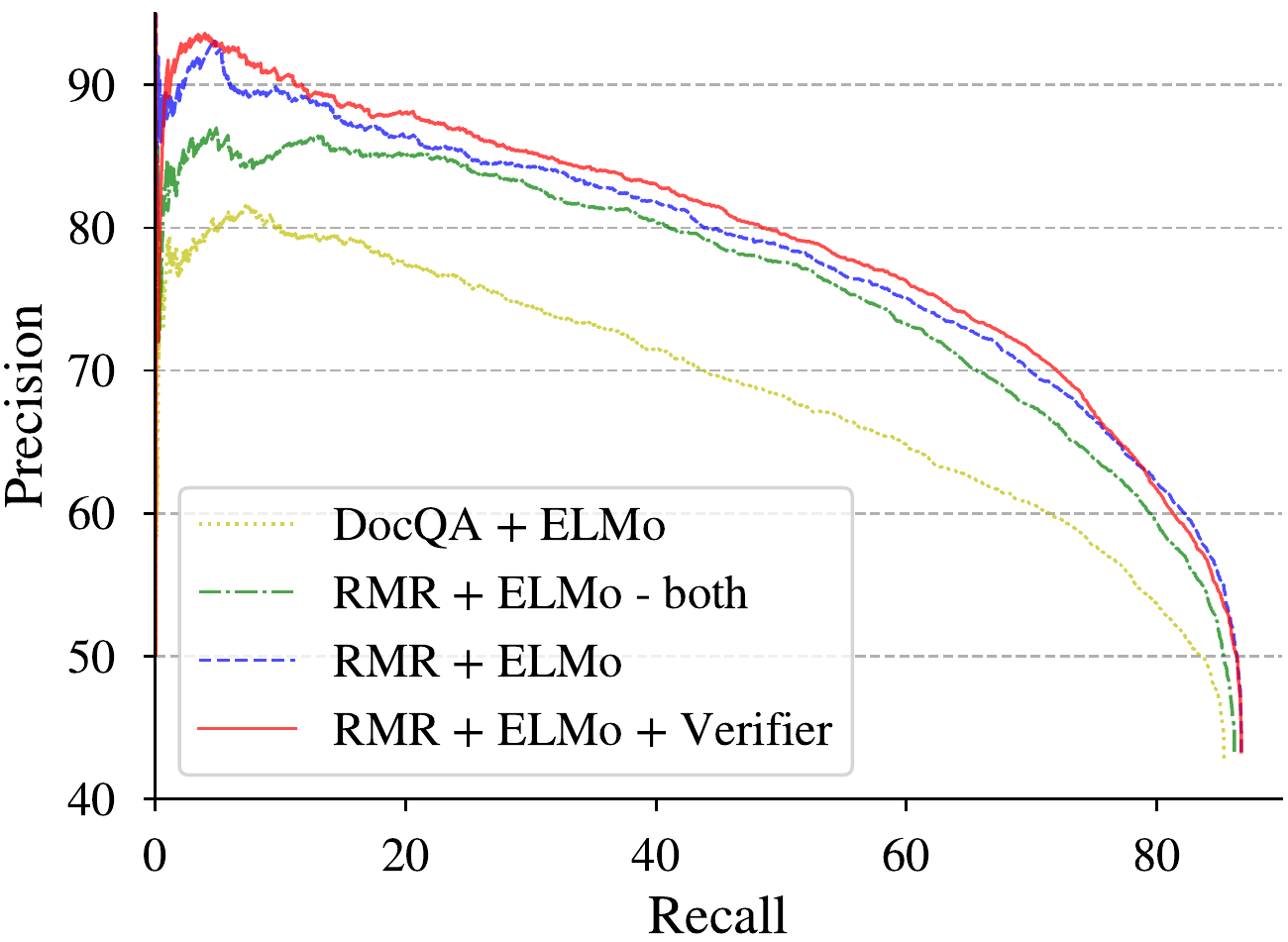}
\vspace{-0.1cm}
\caption{Precision-Recall curves of F1 score.}
\label{fig3}
\vspace{-0.5cm}
\end{center}
\end{figure}

Finally, we plot the precision-recall curves of F1 score on the development set in Figure \ref{fig3}.
We observe that RMR + ELMo + Verifier achieves the best precision when the recall is less than 80.
After the recall exceeds 80, the precision of RMR + ELMo becomes slightly better.
Ablating two auxiliary losses, however, leads to an overall degradation on the curve, but it still outperforms the baseline by a large margin.

\subsection{Error Analysis}

\begin{table}
\begin{center}
\begin{tabular}{l|cccc}
\toprule
\multirow{2}*{ Configuration } & \multicolumn{2}{c}{ All } & NoAns \\
& EM & F1 & ACC \\ 
\midrule
RMR                                                        & 66.9 & 69.1 & 73.1 \\
\ \ + Model-\uppercase\expandafter{\romannumeral1}         & 68.3 & 71.1 & 76.2 \\
\ \ + Model-\uppercase\expandafter{\romannumeral2}         & 68.1 & 70.8 & 75.6  \\
\ \ + Model-\uppercase\expandafter{\romannumeral2} + ELMo  & 68.2 & 70.9 & 75.9  \\
\ \ + Model-\uppercase\expandafter{\romannumeral3}         & \bf{68.5} & \bf{71.5} & \bf{77.1}  \\
\ \ + Model-\uppercase\expandafter{\romannumeral3} + ELMo  & 68.5 & 71.2 & 76.5  \\
\midrule
RMR + ELMo                                                 & 71.4 & 73.7 & 77.0 \\
\ \ + Model-\uppercase\expandafter{\romannumeral1}         & 71.8 & 74.4 & 77.3 \\
\ \ + Model-\uppercase\expandafter{\romannumeral2}         & 71.8 & 74.2 & 78.1 \\
\ \ + Model-\uppercase\expandafter{\romannumeral2} + ELMo  & 72.0 & 74.3 & 78.2 \\
\ \ + Model-\uppercase\expandafter{\romannumeral3}         & \bf{72.3} & \bf{74.8} & \bf{78.6} \\
\ \ + Model-\uppercase\expandafter{\romannumeral3} + ELMo  & 71.8 & 74.3 & 78.3 \\
\bottomrule
\end{tabular}
\caption{\label{table4} Comparison of readers with different answer verifiers.}
\vspace{-0.1cm}
\end{center}
\end{table}

\begin{table}
\begin{center}
\begin{tabular}{l|cccc}
\toprule
\multirow{2}*{ Configuration } & \multicolumn{2}{c}{ All } & NoAns \\
& EM & F1 & ACC \\ 
\midrule
DocQA                                                      & 61.9 & 64.8 & 69.1 \\
\ \ + Model-\uppercase\expandafter{\romannumeral3}         & \bf{66.5} & \bf{69.2} & \bf{75.2}  \\
\midrule
DocQA + ELMo                                               & 65.1 & 67.6 & 70.6 \\
\ \ + Model-\uppercase\expandafter{\romannumeral3}         & \bf{68.0} & \bf{70.7} & \bf{76.1}  \\
\bottomrule
\end{tabular}
\caption{\label{table5} Comparison of different readers with fixed answer verifier.}
\vspace{-0.5cm}
\end{center}
\end{table}

\begin{table*}
\begin{center}
\begin{tabular}{l|ccccc}
\toprule
Configuration           & Case1 \textcolor{red}{\cmark} & Case2 \textcolor{red}{\cmark} & Case3 \textcolor{blue}{\xmark} & Case4 \textcolor{blue}{\xmark} & Case5 \textcolor{blue}{\xmark} \\ 
\midrule
RMR - both              & 27.8\% & 37.3\% & 6.5\% & 12.7\% & 15.7\% \\
RMR	                   & 27\% & \bf{39.9\%} & \bf{5.9\%} & \bf{10.2\%} & 17\% \\
RMR + Verifier          & \bf{30.3\%} & 38.2\% & 8.4\% & 11.8\% & \bf{11.3\%} \\
\midrule
RMR + ELMo - both	   & 31.5\% & 38.3\% & 5.6\% & 11.8\% & 12.8\% \\
RMR + ELMo              & 31.2\% & \bf{40.2\%} & \bf{5.5\%} & \bf{9.9\%} & 13.2\% \\
RMR + ELMo + Verifier   & \bf{32.5\%} & 39.8\% & 6.5\% & 10.3\% & \bf{10.9\%} \\
\bottomrule
\end{tabular}
\caption{\label{table6} Percentage of five categories. Correct predictions are denoted with \textcolor{red}{\cmark}, while wrong cases are marked with \textcolor{blue}{\xmark}.}
\vspace{-0.3cm}
\end{center}
\end{table*}

To perform error analysis, we first categorize all examples on the development set into 5 classes: 
\begin{itemize}
\item \emph{Case1}: the question is \emph{answerable}, the no-answer probability is \emph{less} than the threshold, and the answer is \emph{correct}. 
\item \emph{Case2}: the question is \emph{unanswerable}, and the no-answer probability is \emph{larger} than the threshold.
\item \emph{Case3}: almost the same as case1, except that the predicted answer is \emph{wrong}.
\item \emph{Case4}: the question is \emph{unanswerable}, but the no-answer probability is \emph{less} than the threshold.
\item \emph{Case5}: the question is \emph{answerable}, but the no-answer probability is \emph{larger} than the threshold.
\end{itemize}

We then show the percentage of each category in Table \ref{table6}. 
As we can see, the base reader trained with auxiliary losses is notably better at \emph{case2} and \emph{case4} compared to the baseline, implying that our proposed losses help the model mainly improve upon unanswerable cases.
After adding the answer verifier, we observe that although the system's performance on unanswerable cases slightly decreases, the results on \emph{case1} and \emph{case5} have been improved.
This demonstrates that the answer verifier does well on detecting answerable question rather than unanswerable one.
Besides, we find that the error of answer extraction is relatively small (6.5\% for Case3 in RMR + ELMo + Verifier).
However, the classification error on no-answer detection is much larger. More than 20\% of examples are misclassified even with our best system (10.3\% for Case4 and 10.9\% for Case5 in RMR + ELMo + Verifier).
Therefore, we argue that the main performance bottleneck lies in no-answer detection instead of answer extraction.

Next, to understand the challenges our approach faces, we manually investigate 50 incorrectly predicted unanswerable examples (based on F1) that are randomly sampled from the development set. 
Following the types of negative examples defined by Rajpurkar et al.~\shortcite{Rajpurkar18}, we categorize the sampled examples and show them in Table \ref{table7}.
As we can see, our system is good at recognize \emph{negation} and \emph{antonym}.
The frequency of negation decreases from 9\% to 0\% and only 4 antonym examples are predicted wrongly.
We think that this is because the two types are relatively easier to identify.
Both of negation and antonym only require to detect one single word in the question, such as ``\emph{never}'' or ``\emph{not}'' for negation and ``\emph{increase}'' to ``\emph{decrease}'' for antonym.
However, \emph{impossible condition} and \emph{other neutral} types roughly acount for 46\% of the error set, indicating that our system performs less effectively on these more difficult cases.

\begin{table}
\begin{center}
\begin{tabular}{l|cccc}
\toprule
\multirow{2}*{ Phenomenon } & \multicolumn{2}{c}{ Percentage } \\
& All & Error \\ 
\midrule
Negation  &  9\%  & 0\% \\    
Antonym  &  20\% & 8\% \\    
Entity Swap  & 21\% & 24\% \\    
Mutual Exclusion  & 15\% & 16\% \\    
Impossible Condition  & 4\% & 14\% \\    
Other Neutral  &  24\% & 32\% \\    
Answerable  &  7\% & 6\% \\                                                  
\bottomrule
\end{tabular}
\caption{\label{table7} Linguistic phenomena exhibited by all negative examples (statistics from Rajpurkar et al.~\shortcite{Rajpurkar18}) and sampled error cases of RMR + ELMo + Verifier.}
\vspace{-0.5cm}
\end{center}
\end{table}
\section{Related Work}

\noindent{\bf Reading Comprehension Datasets.} 
Various large-scale reading comprehension datasets, such as cloze-style test~\cite{Hermann15}, answer extraction benchmark~\cite{Rajpurkar16,Joshi17} and answer generation benchmark~\cite{Nguyen16,Kovcisky18}, have been proposed.
However, these datasets still guarantee that the given context must contain an answer.
Recently, some works construct negative examples by retrieving passages for existing questions based on Lucene~\cite{tan2018know} and TF-IDF~\cite{Clark18}, or using crowdworkers to craft unanswerable questions~\cite{Rajpurkar18}.
Compared to automatically retrieved negative examples, human-annotated examples are more difficult to detect for two reasons: (1) the questions are relevant to the passage and (2) the passage contains a plausible answer to the question.
Therefore, we choose to work on the SQuAD 2.0 dataset in this paper.

\noindent{\bf Neural Networks for Reading Comprehension.} 
Neural reading models typically leverage various attention mechanisms to build interdependent representations of passage and question, and sequentially predict the answer boundary~\cite{Seo17,Hu17,Wang17b,Yu18,Hu18}.
However, these approaches are not designed to handle no-answer cases.
To address this problem, previous works~\cite{Levy17,Clark18,kundu2018nil} predict a no-answer probability in addition to the distribution over answer spans, so as to jointly learn no-answer detection as well as answer extraction.
Our no-answer reader extends existing approaches by introducing two auxiliary losses that enhance these two tasks independently.

\noindent{\bf Recognizing Textual Entailment.} 
Recognizing textual entailment (RTE)~\cite{dagan2010recognizing,Marelli14}, or known as natural language inference (NLI)~\cite{Bowman15}, requires systems to understand entailment, contradiction or semantic neutrality between two sentences.
This task is strongly related to no-answer detection, where the machine needs to understand if the passage and the question supports the answer.
To recognize entailment, various branches of works have been proposed, including encoding-based approach~\cite{bowman2016fast,mou2015natural}, interaction-based approach~\cite{Parikh16,chen2016enhanced} and sequence-based approach~\cite{Radford18}.
In this paper we investigate the last two branches and further propose a hybrid architecture that combines both of them properly.

\noindent{\bf Answer Validation.} 
Early answer validation task~\cite{magnini2002right} aims at ranking multiple candidate answers to return a most reliable one.
Later, the answer validation exercise~\cite{rodrigo2008overview} has been proposed to decide whether an answer is correct or not according to a given supporting text and a question, but the dataset is too small for neural network-based approaches. 
Recently, Tan et al.~\shortcite{tan2018know} propose to validate the candidate answer for detecting unanswerable questions, by comparing the question with the passage.
Our answer verifier, on the contrary, denoises the passage by comparing questions with answer sentences, so as to focus on finding local entailment that supports the answer. 
\section{Conclusion}
We proposed a read-then-verify system that is able to abstain from answering when a question has no answer given the passage.
We first introduce two auxiliary losses to help the reader concentrate on answer extraction and no-answer detection respectively, and then utilize an answer verifier to validate the legitimacy of the predicted answer, in which three different architectures are investigated.
Our system has achieved state-of-the-art results on the SQuAD 2.0 dataset at the time of submission (Aug. 28th, 2018).
Looking forward, we plan to design new structures for answer verifiers to handle questions with more complicated inferences.

\section*{Acknowledgments}
We would like to thank Pranav Rajpurkar and Robin Jia for their helps with SQuAD 2.0 submissions. This work is supported by the Major State Research Development Program (2016YFB0201305).

\bibliography{sections/reference}

\begin{thebibliography}{}

\bibitem[\protect\citeauthoryear{Bowman \bgroup et al\mbox.\egroup
  }{2015}]{Bowman15}
Bowman, S.~R.; Angeli, G.; Potts, C.; and Manning, C.~D.
\newblock 2015.
\newblock A large annotated corpus for learning natural language inference.
\newblock In {\em Proceedings of EMNLP}.

\bibitem[\protect\citeauthoryear{Bowman \bgroup et al\mbox.\egroup
  }{2016}]{bowman2016fast}
Bowman, S.~R.; Gauthier, J.; Rastogi, A.; Gupta, R.; Manning, C.~D.; and Potts,
  C.
\newblock 2016.
\newblock A fast unified model for parsing and sentence understanding.
\newblock {\em arXiv preprint arXiv:1603.06021}.

\bibitem[\protect\citeauthoryear{Chen \bgroup et al\mbox.\egroup
  }{2016}]{chen2016enhanced}
Chen, Q.; Zhu, X.; Ling, Z.; Wei, S.; Jiang, H.; and Inkpen, D.
\newblock 2016.
\newblock Enhanced lstm for natural language inference.
\newblock {\em arXiv preprint arXiv:1609.06038}.

\bibitem[\protect\citeauthoryear{Clark and Gardner}{2018}]{Clark18}
Clark, C., and Gardner, M.
\newblock 2018.
\newblock Simple and effective multi-paragraph reading comprehension.
\newblock In {\em Proceedings of ACL}.

\bibitem[\protect\citeauthoryear{Dagan \bgroup et al\mbox.\egroup
  }{2010}]{dagan2010recognizing}
Dagan, I.; Dolan, B.; Magnini, B.; and Roth, D.
\newblock 2010.
\newblock Recognizing textual entailment: rational, evaluation and approaches.
\newblock {\em Natural Language Engineering} 16(1):105--105.

\bibitem[\protect\citeauthoryear{Hendrycks and Gimpel}{2016}]{Hendrycks16}
Hendrycks, D., and Gimpel, K.
\newblock 2016.
\newblock Bridging nonlinearities and stochastic regularizers with gaussian
  error linear units.
\newblock {\em arXiv preprint arXiv:1606.08415}.

\bibitem[\protect\citeauthoryear{Hermann \bgroup et al\mbox.\egroup
  }{2015}]{Hermann15}
Hermann, K.~M.; Kocisky, T.; Grefenstette, E.; Espeholt, L.; Kay, W.; Suleyman,
  M.; and Blunsom, P.
\newblock 2015.
\newblock Teaching machines to read and comprehend.
\newblock In {\em Proceedings of NIPS}.

\bibitem[\protect\citeauthoryear{Hochreiter and
  Schmidhuber}{1997}]{Hochreiter97}
Hochreiter, S., and Schmidhuber, J.
\newblock 1997.
\newblock Long short-term memory.
\newblock {\em Neural computation} 9(8):1735–--1780.

\bibitem[\protect\citeauthoryear{Hu \bgroup et al\mbox.\egroup }{2018a}]{Hu17}
Hu, M.; Peng, Y.; Huang, Z.; Qiu, X.; Wei, F.; and Zhou, M.
\newblock 2018a.
\newblock Reinforced mnemonic reader for machine reading comprehension.
\newblock In {\em Proceedings of IJCAI}.

\bibitem[\protect\citeauthoryear{Hu \bgroup et al\mbox.\egroup }{2018b}]{Hu18}
Hu, M.; Peng, Y.; Wei, F.; Huang, Z.; DongshengLi; Yang, N.; and Zhou, M.
\newblock 2018b.
\newblock Attention-guided answer distillation for machine reading
  comprehension.
\newblock In {\em Proceedings of EMNLP}.

\bibitem[\protect\citeauthoryear{Huang \bgroup et al\mbox.\egroup
  }{2018}]{Huang17b}
Huang, H.-Y.; Zhu, C.; Shen, Y.; and Chen, W.
\newblock 2018.
\newblock Fusionnet: fusing via fully-aware attention with application to
  machine comprehension.
\newblock In {\em Proceedings of ICLR}.

\bibitem[\protect\citeauthoryear{Joshi \bgroup et al\mbox.\egroup
  }{2017}]{Joshi17}
Joshi, M.; Choi, E.; Weld, D.~S.; and Zettlemoyer, L.
\newblock 2017.
\newblock Triviaqa: a large scale distantly supervised challenge dataset for
  reading comprehension.
\newblock In {\em Proceedings of ACL}.

\bibitem[\protect\citeauthoryear{Kingma and Ba}{2014}]{Kingma14}
Kingma, D.~P., and Ba, L.~J.
\newblock 2014.
\newblock Adam: A method for stochastic optimization.
\newblock In {\em CoRR, abs/1412.6980}.

\bibitem[\protect\citeauthoryear{Ko{\v{c}}isk{\`y} \bgroup et al\mbox.\egroup
  }{2018}]{Kovcisky18}
Ko{\v{c}}isk{\`y}, T.; Schwarz, J.; Blunsom, P.; Dyer, C.; Hermann, K.~M.;
  Melis, G.; and Grefenstette, E.
\newblock 2018.
\newblock The narrativeqa reading comprehension challenge.
\newblock {\em Transactions of ACL} 6:317--328.

\bibitem[\protect\citeauthoryear{Kundu and Ng}{2018}]{kundu2018nil}
Kundu, S., and Ng, H.~T.
\newblock 2018.
\newblock A nil-aware answer extraction framework for question answering.
\newblock In {\em Proceedings of EMNLP},  4243--4252.

\bibitem[\protect\citeauthoryear{Levy \bgroup et al\mbox.\egroup
  }{2017}]{Levy17}
Levy, O.; Seo, M.; Choi, E.; and Zettlemoyer, L.
\newblock 2017.
\newblock Zero-shot relation extraction via reading comprehension.
\newblock {\em arXiv preprint arXiv:1706.04115}.

\bibitem[\protect\citeauthoryear{Liu \bgroup et al\mbox.\egroup
  }{2018a}]{Liu18}
Liu, P.~J.; Saleh, M.; Pot, E.; Goodrich, B.; Sepassi, R.; Kaiser, L.; and
  Shazeer, N.
\newblock 2018a.
\newblock Generating wikipedia by summarizing long sequences.
\newblock {\em arXiv preprint arXiv:1801.10198}.

\bibitem[\protect\citeauthoryear{Liu \bgroup et al\mbox.\egroup
  }{2018b}]{liu2017stochastic}
Liu, X.; Shen, Y.; Duh, K.; and Gao, J.
\newblock 2018b.
\newblock Stochastic answer networks for machine reading comprehension.
\newblock In {\em Proceedings of ACL}.

\bibitem[\protect\citeauthoryear{Magnini \bgroup et al\mbox.\egroup
  }{2002}]{magnini2002right}
Magnini, B.; Negri, M.; Prevete, R.; and Tanev, H.
\newblock 2002.
\newblock Is it the right answer? exploiting web redundancy for answer
  validation.
\newblock In {\em Proceedings of ACL}.

\bibitem[\protect\citeauthoryear{Marelli \bgroup et al\mbox.\egroup
  }{2014}]{Marelli14}
Marelli, M.; Menini, S.; Baroni, M.; Bentivogli, L.; Bernardi, R.; Zamparelli,
  R.; et~al.
\newblock 2014.
\newblock A sick cure for the evaluation of compositional distributional
  semantic models.
\newblock In {\em LREC},  216--223.

\bibitem[\protect\citeauthoryear{Mou \bgroup et al\mbox.\egroup
  }{2015}]{mou2015natural}
Mou, L.; Men, R.; Li, G.; Xu, Y.; Zhang, L.; Yan, R.; and Jin, Z.
\newblock 2015.
\newblock Natural language inference by tree-based convolution and heuristic
  matching.
\newblock {\em arXiv preprint arXiv:1512.08422}.

\bibitem[\protect\citeauthoryear{Nguyen \bgroup et al\mbox.\egroup
  }{2016}]{Nguyen16}
Nguyen, T.; Rosenberg, M.; Song, X.; Gao, J.; Tiwary, S.; Majumder, R.; and
  Deng, L.
\newblock 2016.
\newblock Ms marco: a human generated machine reading comprehension dataset.
\newblock {\em arXiv preprint arXiv:1611.09268}.

\bibitem[\protect\citeauthoryear{Parikh \bgroup et al\mbox.\egroup
  }{2016}]{Parikh16}
Parikh, A.~P.; T{\"a}ckstr{\"o}m, O.; Das, D.; and Uszkoreit, J.
\newblock 2016.
\newblock A decomposable attention model for natural language inference.
\newblock {\em arXiv preprint arXiv:1606.01933}.

\bibitem[\protect\citeauthoryear{Pennington, Socher, and
  Manning}{2014}]{Pennington14}
Pennington, J.; Socher, R.; and Manning, C.~D.
\newblock 2014.
\newblock Glove: Global vectors for word representation.
\newblock In {\em Proceedings of EMNLP}.

\bibitem[\protect\citeauthoryear{Peters \bgroup et al\mbox.\egroup
  }{2018}]{Elmo17}
Peters, M.~E.; Neumann, M.; Iyyer, M.; Gardner, M.; Clark, C.; Lee, K.; and
  Zettlemoyer, L.
\newblock 2018.
\newblock Deep contextualized word prepresentations.
\newblock In {\em Proceedings of NAACL}.

\bibitem[\protect\citeauthoryear{Radford \bgroup et al\mbox.\egroup
  }{2018}]{Radford18}
Radford, A.; Narasimhan, K.; Salimans, T.; and Sutskever, I.
\newblock 2018.
\newblock Improving language understanding by generative pre-training.

\bibitem[\protect\citeauthoryear{Rajpurkar \bgroup et al\mbox.\egroup
  }{2016}]{Rajpurkar16}
Rajpurkar, P.; Zhang, J.; Lopyrev, K.; and Liang, P.
\newblock 2016.
\newblock Squad: 100,000+ questions for machine comprehension of text.
\newblock In {\em Proceedings of EMNLP}.

\bibitem[\protect\citeauthoryear{Rajpurkar, Jia, and Liang}{2018}]{Rajpurkar18}
Rajpurkar, P.; Jia, R.; and Liang, P.
\newblock 2018.
\newblock Know what you don't know: unanswerable questions for squad.
\newblock In {\em Proceedings of ACL}.

\bibitem[\protect\citeauthoryear{Rodrigo, Pe{\~n}as, and
  Verdejo}{2008}]{rodrigo2008overview}
Rodrigo, {\'A}.; Pe{\~n}as, A.; and Verdejo, F.
\newblock 2008.
\newblock Overview of the answer validation exercise 2008.
\newblock In {\em Workshop of CLEF},  296--313.
\newblock Springer.

\bibitem[\protect\citeauthoryear{Sennrich, Haddow, and
  Birch}{2016}]{Sennrich16}
Sennrich, R.; Haddow, B.; and Birch, A.
\newblock 2016.
\newblock Neural machine translation of rare words with subword units.
\newblock In {\em Proceedings of ACL}.

\bibitem[\protect\citeauthoryear{Seo \bgroup et al\mbox.\egroup }{2017}]{Seo17}
Seo, M.; Kembhavi, A.; Farhadi, A.; and Hajishirzi, H.
\newblock 2017.
\newblock Bidirectional attention flow for machine comprehension.
\newblock In {\em Proceedings of ICLR}.

\bibitem[\protect\citeauthoryear{Srivastava \bgroup et al\mbox.\egroup
  }{2014}]{Srivastava14}
Srivastava, N.; Hinton, G.; Krizhevsky, A.; Sutskever, I.; and Salakhutdinov,
  R.
\newblock 2014.
\newblock Dropout: a simple way to prevent neural networks from overfitting.
\newblock {\em JMLR}  1929--1958.

\bibitem[\protect\citeauthoryear{Tan \bgroup et al\mbox.\egroup
  }{2018}]{tan2018know}
Tan, C.; Wei, F.; Zhou, Q.; Yang, N.; Lv, W.; and Zhou, M.
\newblock 2018.
\newblock I know there is no answer: modeling answer validation for machine
  reading comprehension.
\newblock In {\em Proceedings of NLPCC},  85--97.
\newblock Springer.

\bibitem[\protect\citeauthoryear{Vaswani \bgroup et al\mbox.\egroup
  }{2017}]{vaswani2017attention}
Vaswani, A.; Shazeer, N.; Parmar, N.; Uszkoreit, J.; Jones, L.; Gomez, A.~N.;
  Kaiser, {\L}.; and Polosukhin, I.
\newblock 2017.
\newblock Attention is all you need.
\newblock In {\em Proceedings of NIPS},  5998--6008.

\bibitem[\protect\citeauthoryear{Vinyals, Fortunato, and
  Jaitly}{2015}]{Vinyals15}
Vinyals, O.; Fortunato, M.; and Jaitly, N.
\newblock 2015.
\newblock Pointer networks.
\newblock In {\em Proceedings of NIPS}.

\bibitem[\protect\citeauthoryear{Wang \bgroup et al\mbox.\egroup
  }{2017}]{Wang17b}
Wang, W.; Yang, N.; Wei, F.; Chang, B.; and Zhou, M.
\newblock 2017.
\newblock Gated self-matching networks for reading comprehension and question
  answering.
\newblock In {\em Proceedings of ACL}.

\bibitem[\protect\citeauthoryear{Wang, Yan, and Wu}{2018}]{wang2018multi}
Wang, W.; Yan, M.; and Wu, C.
\newblock 2018.
\newblock Multi-granularity hierarchical attention fusion networks for reading
  comprehension and question answering.
\newblock In {\em Proceedings of ACL}.

\bibitem[\protect\citeauthoryear{Yu \bgroup et al\mbox.\egroup }{2018}]{Yu18}
Yu, A.~W.; Dohan, D.; Luong, M.-T.; Zhao, R.; Chen, K.; Norouzi, M.; and Le,
  Q.~V.
\newblock 2018.
\newblock Qanet: combining local convolution with global self-attention for
  reading comprehension.
\newblock In {\em Proceedings of ICLR}.

\end{thebibliography}
\bibliographystyle{aaai}

\end{document}